\documentclass{article}
\usepackage{times}
\usepackage{float}
\usepackage{helvet}
\usepackage{courier}
\usepackage[utf8]{inputenc} 
\usepackage[T1]{fontenc}    
\usepackage[font={small}]{caption}
\usepackage{url}            
\usepackage{booktabs}       
\usepackage{amsfonts}       
\usepackage{nicefrac}       
\usepackage{microtype}      
\usepackage{xcolor}         
\usepackage{amsmath}
\usepackage{xspace}
\usepackage{cleveref}
\usepackage{algorithm}
\usepackage{algpseudocode}
\usepackage{graphicx} 
\usepackage{indentfirst}
\usepackage{subcaption}
\usepackage{todonotes}
\usepackage{pgfplots}
\usepackage{colortbl}
\usepackage{tabularx}
\usepackage{xcolor}
\usepackage{multicol}
\usepackage{multirow}
\graphicspath{ {Figures/} }

\newcommand{\mathdefault}[1][]{}

\newcommand{\kosmos}{\textsc{KOSMOS}\xspace}
\newcommand{\flux}{\textsc{Flux}\xspace}
\newcommand{\dino}{\textsc{GroundingDINO}\xspace}
\newcommand{\detr}{\textsc{DETR}\xspace}
\newcommand{\spacy}{\textsc{spaCy}\xspace}
\newcommand{\scipy}{\textsc{SciPy}\xspace}
\newcommand{\sd}{\textsc{Stable-Diffusion 3}\xspace}
\newcommand{\kand}{\textsc{Kandinsky 2.2}\xspace}

\newcommand{\rs}{\textit{requirement satisfaction}\xspace}
\newcommand{\RS}{\textit{Satisfaction of Prompt Requirements}\xspace}
\newcommand{\dive}{\textit{diversity}\xspace}
\newcommand{\Dive}{\textit{Diversity}\xspace}
\newcommand{\coh}{\textit{cohesion}\xspace}
\newcommand{\Coh}{\textit{Cohesion}\xspace}
\newcommand{\creativity}{\textit{creativity ranking}\xspace}

\newcommand{\cohfact}{\textit{cohesion factor}\xspace}
\newcommand{\CohFact}{\textit{Cohesion Factor}\xspace}

\newcommand{\divefact}{\textit{diversity factor}\xspace}
\newcommand{\DiveFact}{\textit{Diversity Factor}\xspace}

\newcommand{\imgtoimg}{\texttt{img2img}\space}
\newcommand{\texttoimg}{\texttt{text2img}\space}

\newcommand{\imonly}{\texttt{img\_only}}
\newcommand{\caponly}{\texttt{cap\_only}}
\newcommand{\imcap}{\texttt{img\_cap}}

\newcommand{\satfact}{\textit{satisfaction factor}\xspace}
\newcommand{\SatFact}{\textit{Requirements Satisfaction Factor}\xspace}

\newtheorem{lemma}{Lemma}
\newtheorem{definition}{Definition}
\newtheorem{proposition}{Proposition}

\newcommand{\lem}{\begin{lemma}}
\newcommand{\elem}{\end{lemma}}
\newcommand{\pro}{\begin{proposition}}
\newcommand{\epro}{\end{proposition}}
\newcommand{\dfn}{\begin{definition}}
\newcommand{\edfn}{\end{definition}}

\pgfplotsset{compat=1.18}

\newcommand{\commentout}[1]{}

\title{Quantitative Measures of Task-Oriented Creativity in Popular  Generative Vision Models}
\author{
    Aditi Ramaswamy\\
    \texttt{aditi.ramaswamy@kcl.ac.uk}\\
    \and
    \textbf{Hana Chockler}\\
    \texttt{hana.chockler@kcl.ac.uk}\\
    \and
    \textbf{Melane Navaratnarajah}\\
    \texttt{melane.navaratnarajah@kcl.ac.uk}
}

\begin{document} 
\pagenumbering{arabic}
\maketitle

\begin{abstract}
\begin{quote}
Creativity of generative AI models has been a subject of scientific debate in the last years, without a conclusive answer. In this paper,
we study creativity from a practical perspective and introduce quantitative measures that help the user to choose a suitable
AI model for a given task. We evaluated our measures on a number of popular \imgtoimg generation models and the results suggest
that our measures conform to the intuition.
\end{quote}
\end{abstract}

\begin{figure*}[!htb]
    \small
    \begin{subfigure}{\textwidth}
    \centering
    \includegraphics[width=0.08\textwidth]{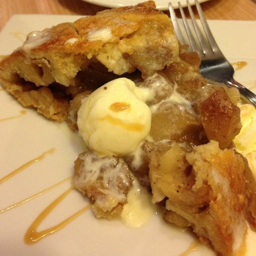}
    \includegraphics[width=0.08\textwidth]{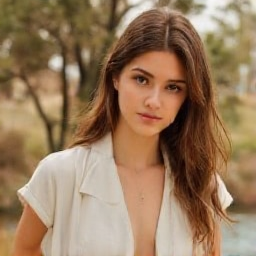}
    \includegraphics[width=0.08\textwidth]{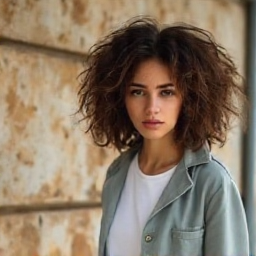}
    \includegraphics[width=0.08\textwidth]{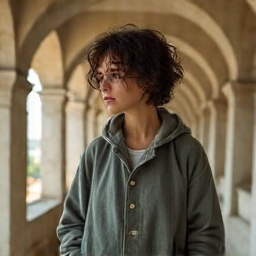}
    \includegraphics[width=0.08\textwidth]{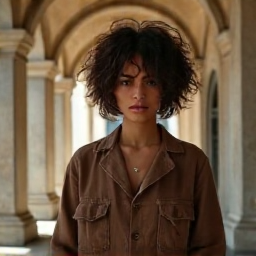}
    \includegraphics[width=0.08\textwidth]{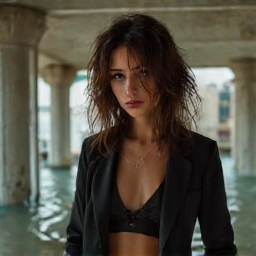}
    \includegraphics[width=0.08\textwidth]{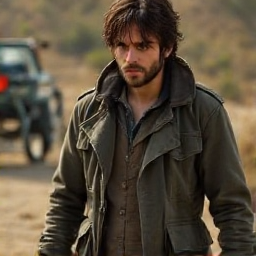}
    \includegraphics[width=0.08\textwidth]{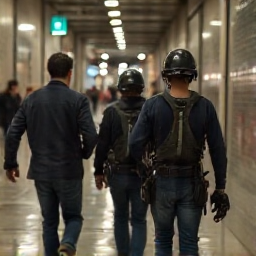}
    \includegraphics[width=0.08\textwidth]{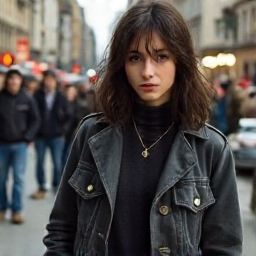}
    \includegraphics[width=0.08\textwidth]{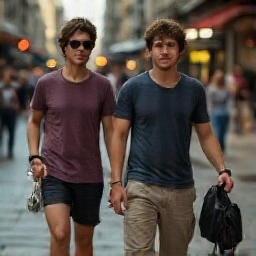}
    \includegraphics[width=0.08\textwidth]{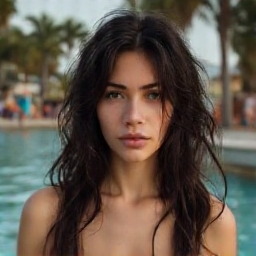}
    \end{subfigure}

    \begin{subfigure}{\textwidth}
    \centering
    \includegraphics[width=0.08\textwidth]{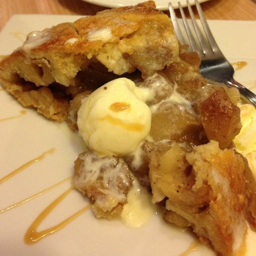}
    \includegraphics[width=0.08\textwidth]{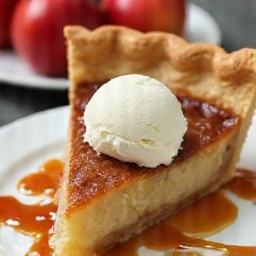}
    \includegraphics[width=0.08\textwidth]{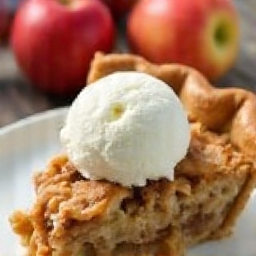}
    \includegraphics[width=0.08\textwidth]{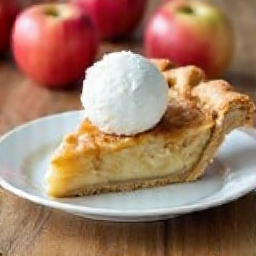}
    \includegraphics[width=0.08\textwidth]{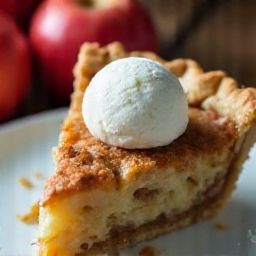}
    \includegraphics[width=0.08\textwidth]{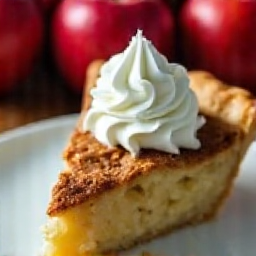}
    \includegraphics[width=0.08\textwidth]{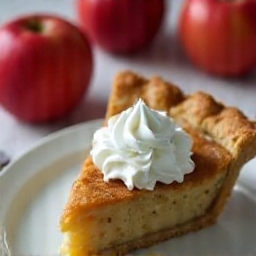}
    \includegraphics[width=0.08\textwidth]{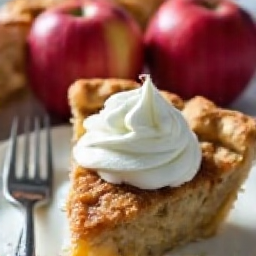}
    \includegraphics[width=0.08\textwidth]{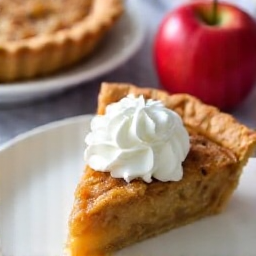}
    \includegraphics[width=0.08\textwidth]{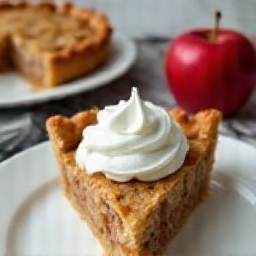}
    \includegraphics[width=0.08\textwidth]{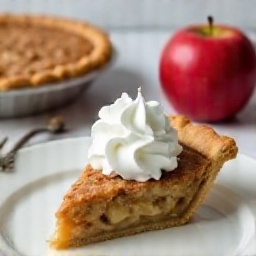}
    \end{subfigure}

    \begin{subfigure}{\textwidth}
    \centering
    \includegraphics[width=0.08\textwidth]{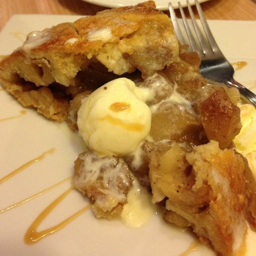}
    \includegraphics[width=0.08\textwidth]{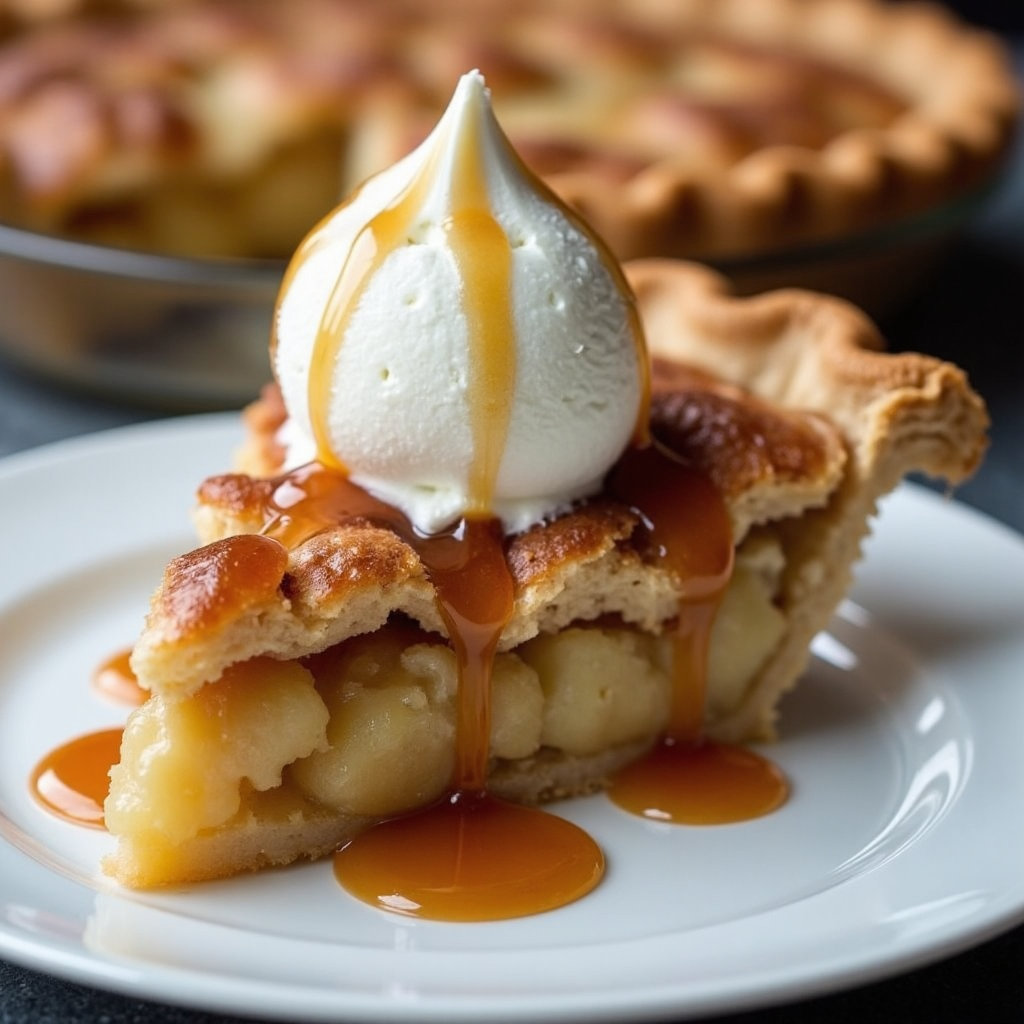}
    \includegraphics[width=0.08\textwidth]{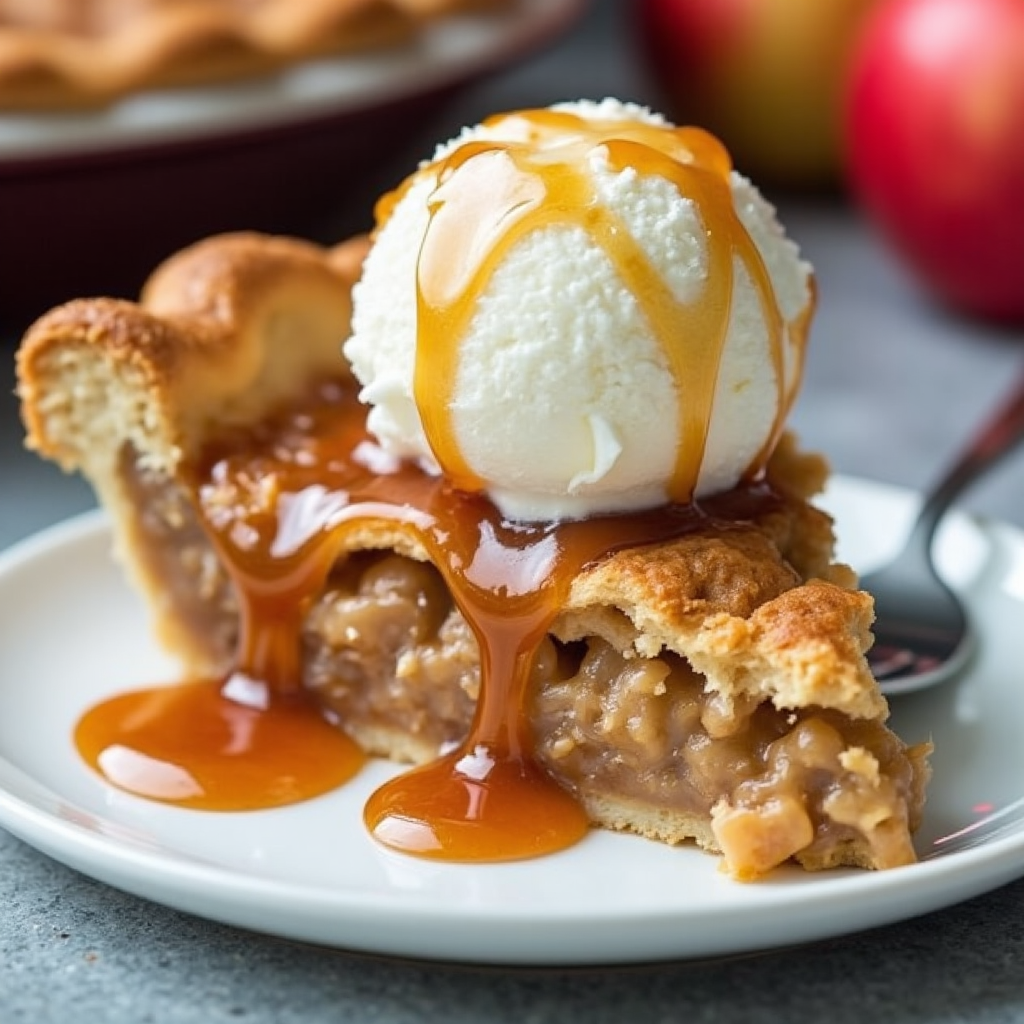}
    \includegraphics[width=0.08\textwidth]{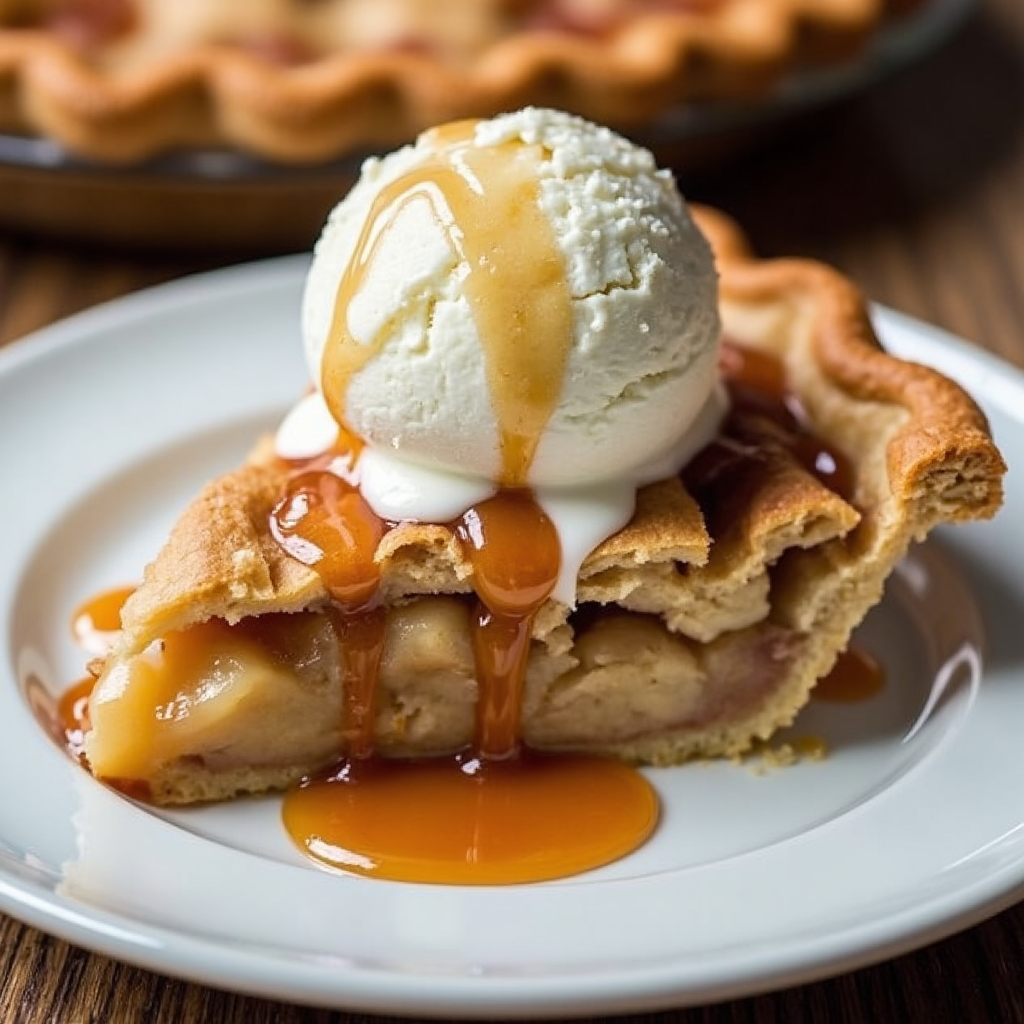}
    \includegraphics[width=0.08\textwidth]{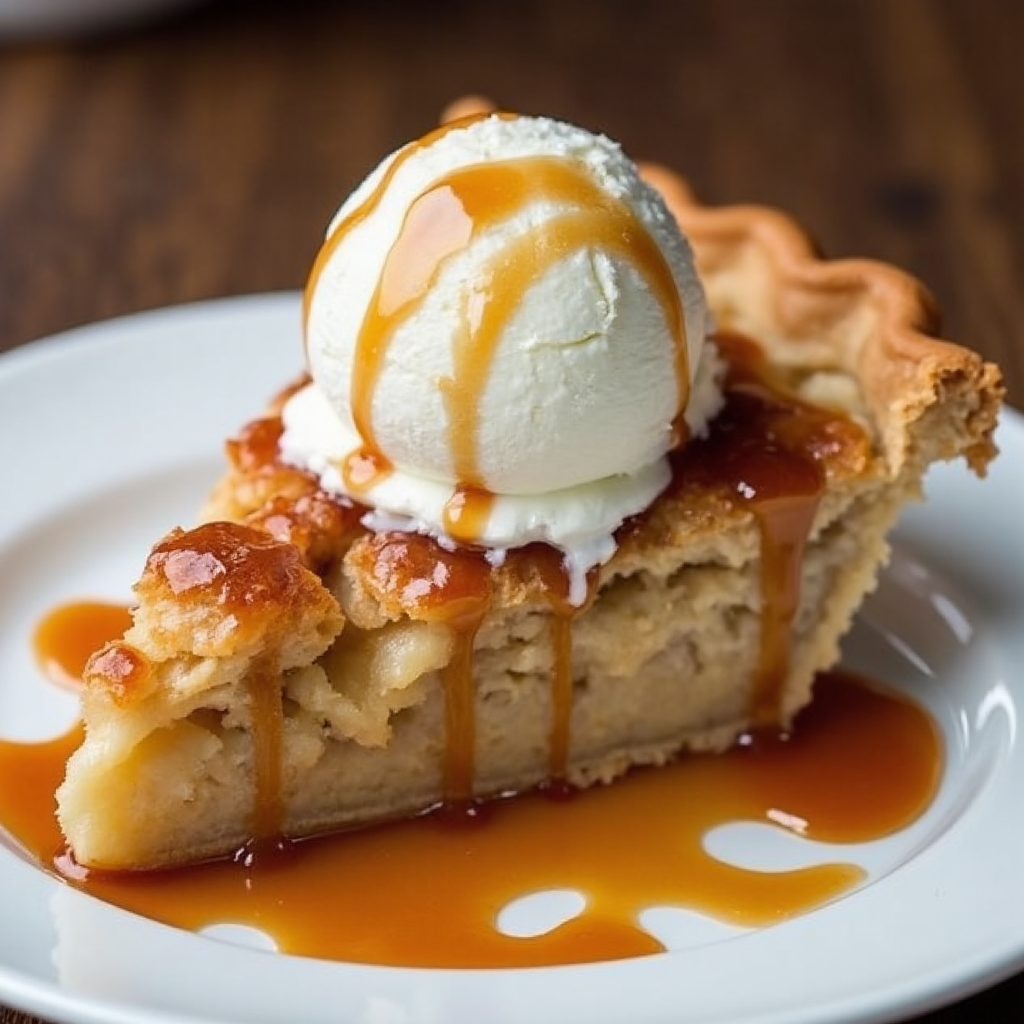}
    \includegraphics[width=0.08\textwidth]{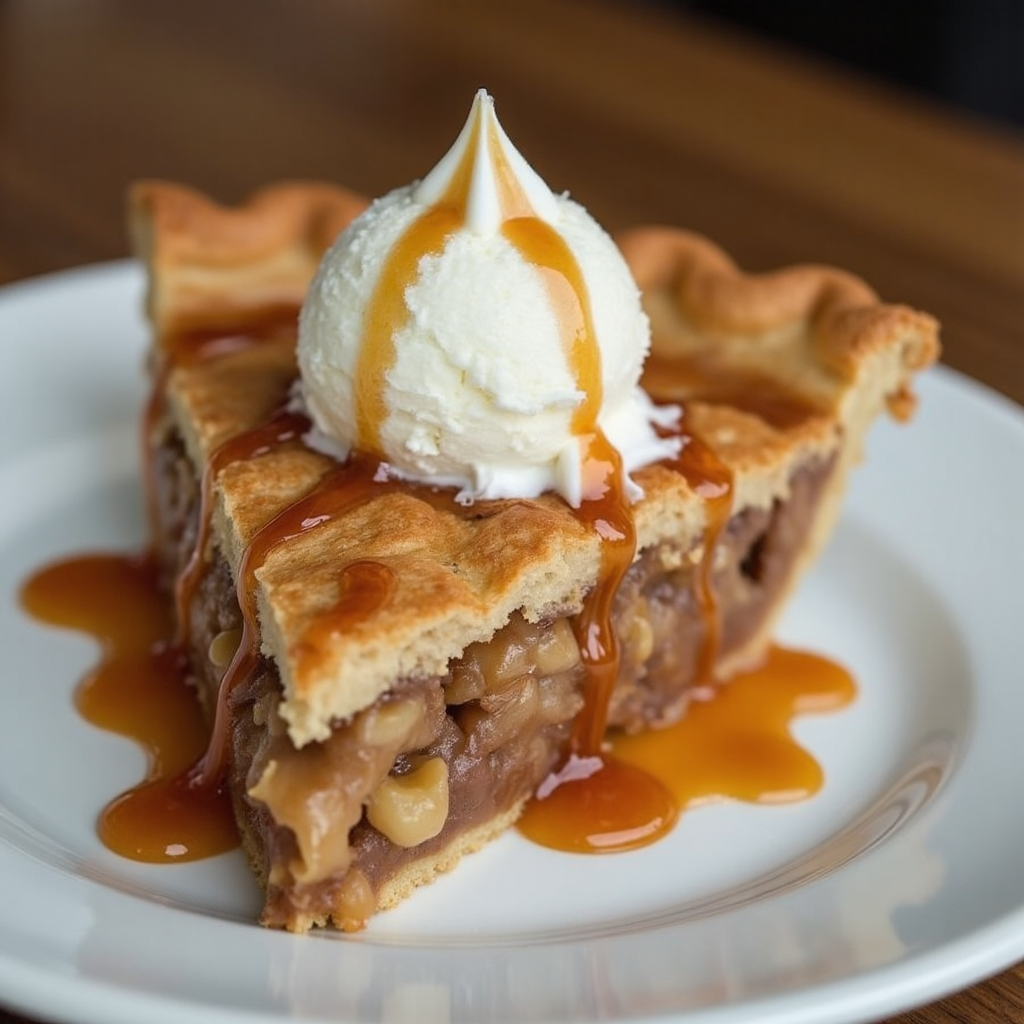}
    \includegraphics[width=0.08\textwidth]{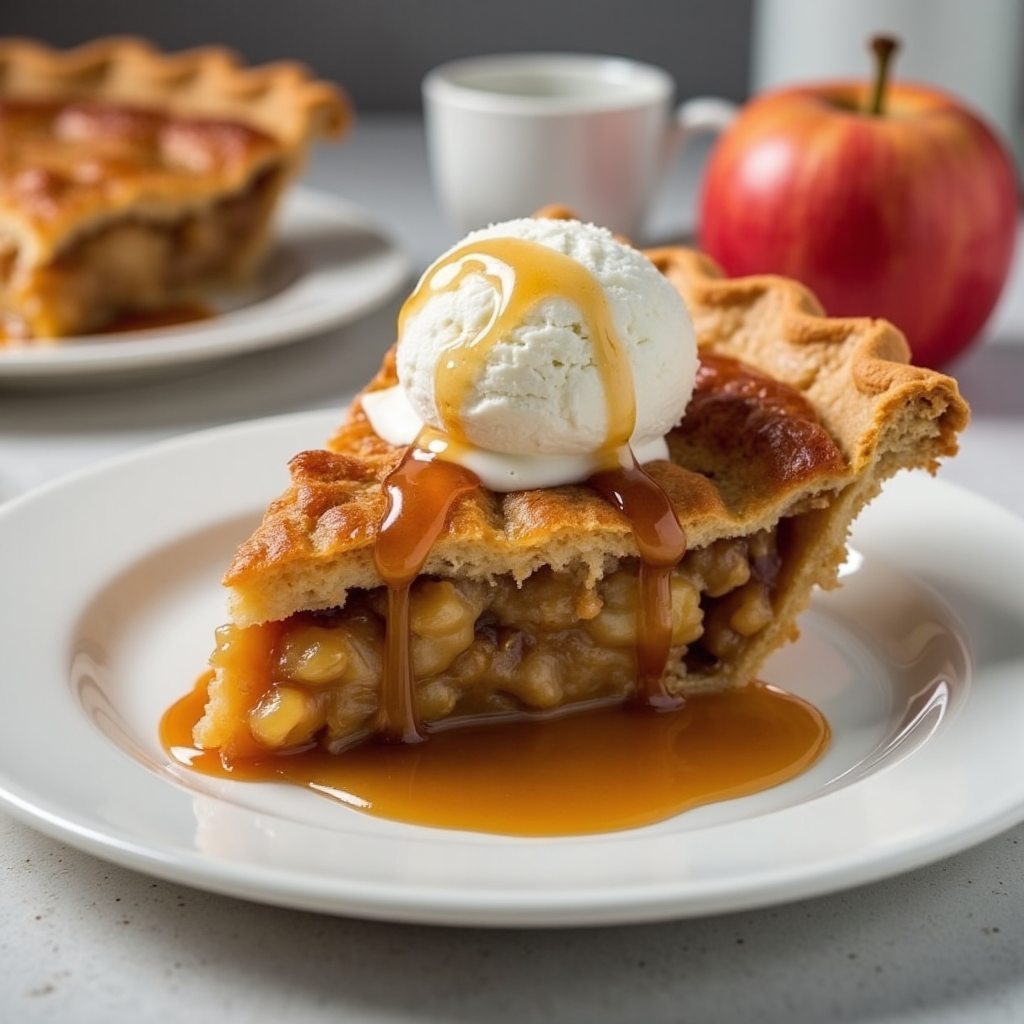}
    \includegraphics[width=0.08\textwidth]{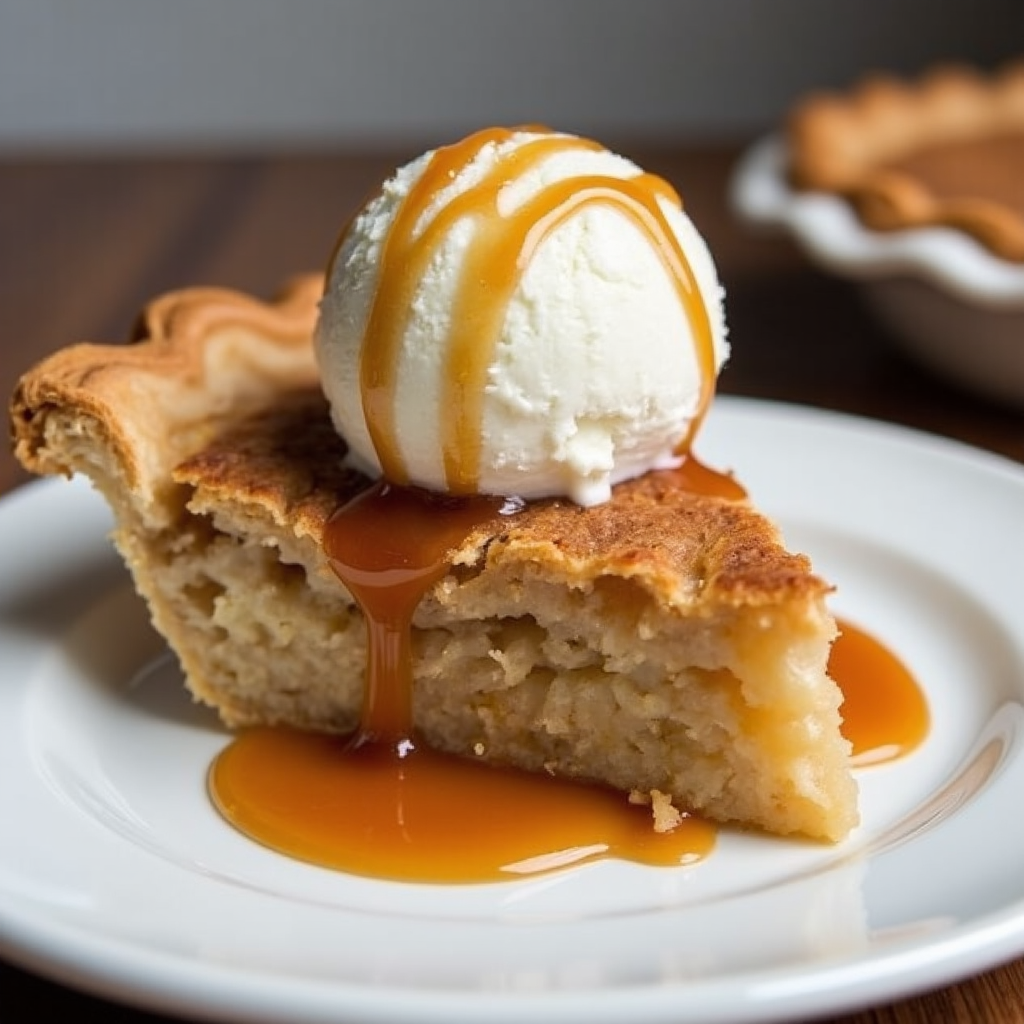}
    \includegraphics[width=0.08\textwidth]{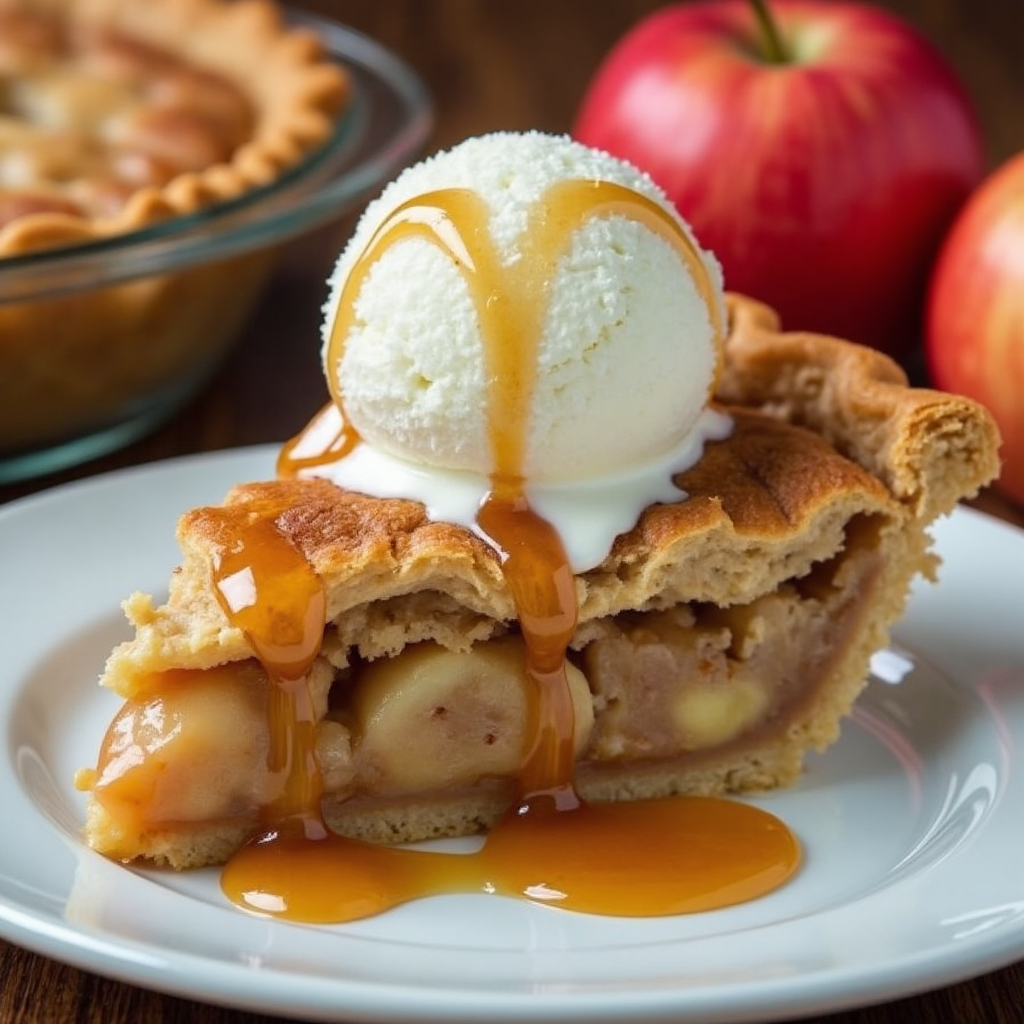}
    \includegraphics[width=0.08\textwidth]{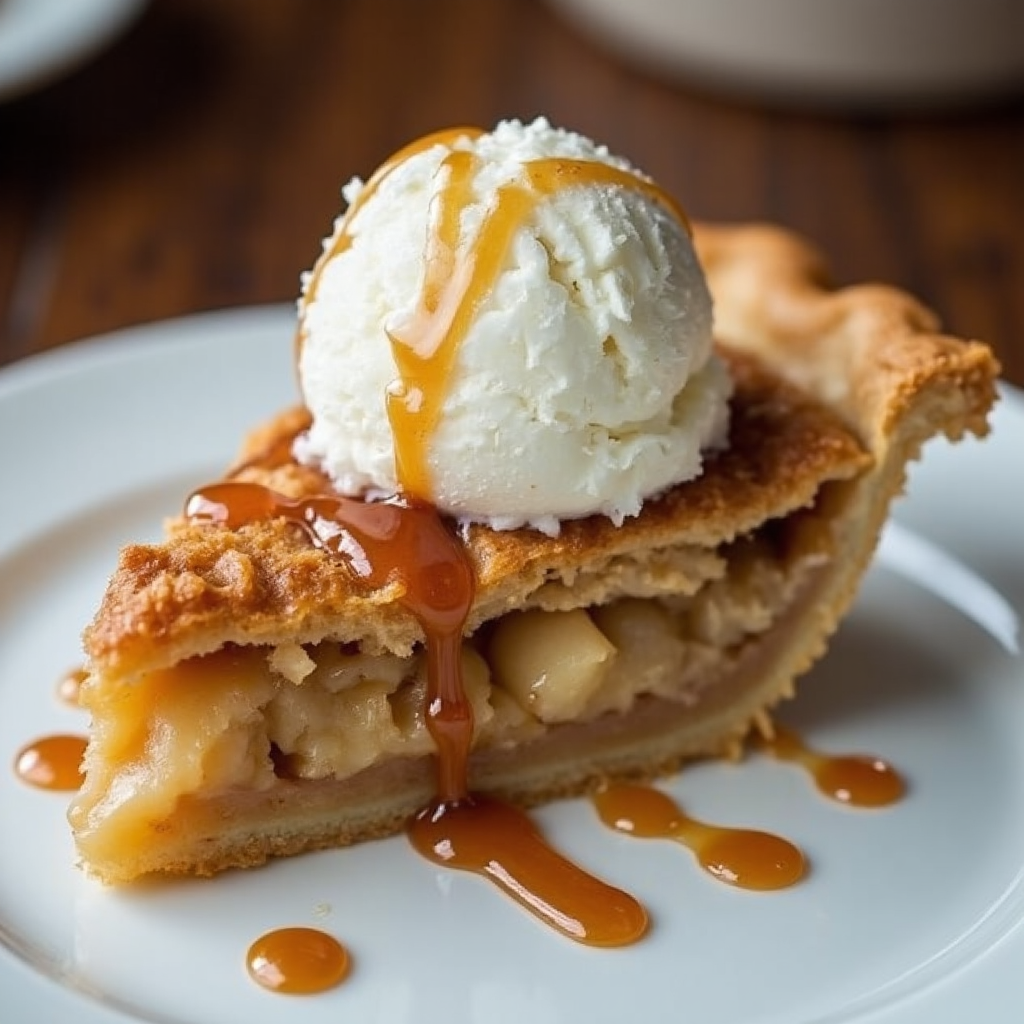}
    \includegraphics[width=0.08\textwidth]{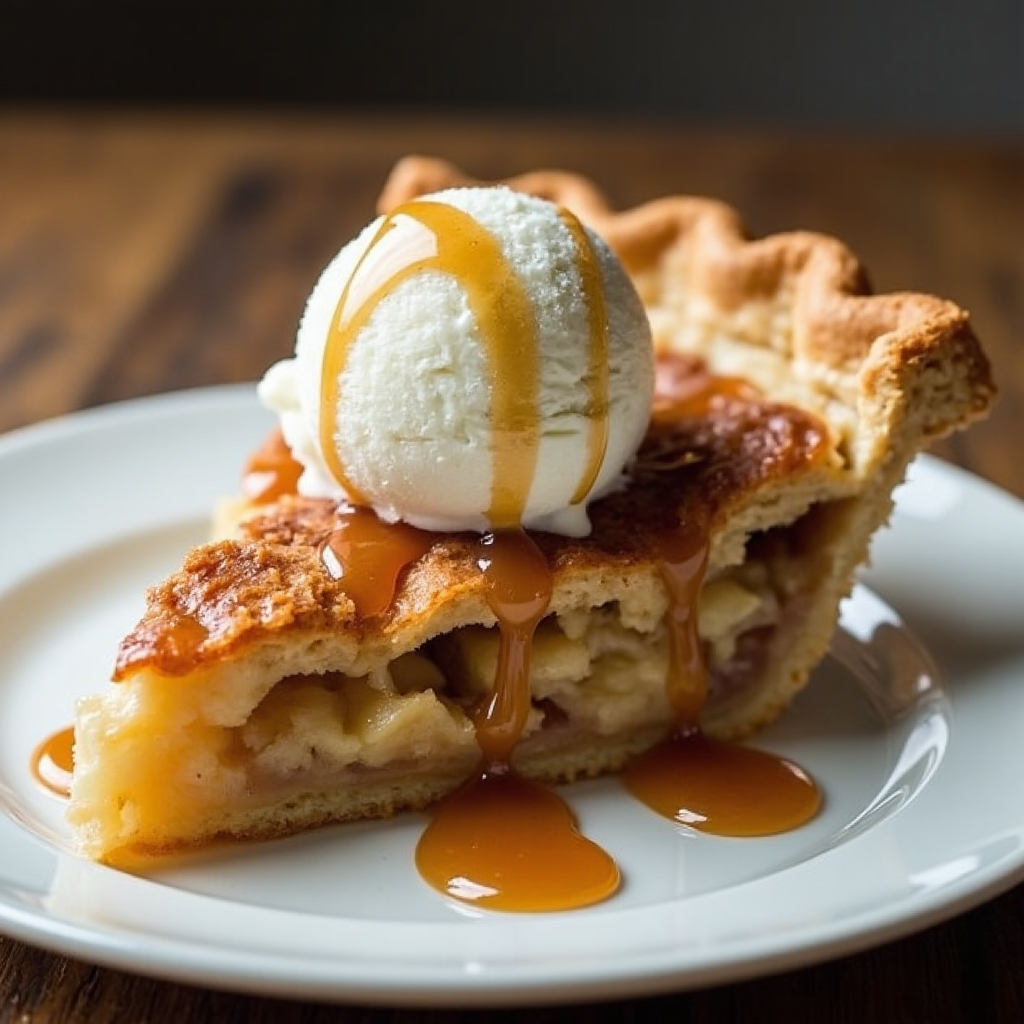}
    \end{subfigure}
    \caption{Example \flux chains, generated from a single seed image \texttt{apple\_pie\_139.png} (the leftmost image), 
    with different input temperatures.}
    \label{fig:flux-strength09-chain-types-examples}
\end{figure*}

\section{Introduction}

Generative AI has become the center of a series of heated debates, one of the biggest ones centered around the concept of human creativity. Some researchers argue that creativity is derived from an innately human ``sociocultural context'', thus excluding AI models \cite{Oppenlaender_2022,Kaufman_2019,Wingstrom_2022}. However, other definitions focus on purely behavioral requirements, such as ``domain-relevant skills'', ``creativity-relevant processes'', and ``extrinsic motivation'', all of which can be exhibited by generative AI models \cite{Amabile_1983}. Related to this, Margaret Boden proposes an AI-relevant cognitive dimension of creativity separate from emotion, focusing on behaviors such as novelty and association of ideas \cite{Boden_1998}. Boden also discusses ``exploratory'' creativity, the ``generation of novel ideas'' within a space, such as an AI model responding to a given task by creating new content within the parameters of the task. These process-oriented definitions suggest that some generative models can exhibit a form of \emph{task-oriented} creative behavior in response to prompts, thus forming the motivation for this paper.

Quantifying this task-oriented creative behavior could provide valuable insights for both the users who interact with these models and the model developers. If a user is presented with clear information about the ways in which a given generative model may exhibit creative behavior in response to their queries, they can make an informed choice about whether they wish to use it. The same information could also pinpoint specific areas of change which those who develop generative models can use to tweak their training data and model architecture to better suit their intended purpose.

The focus of this paper is answering the following
research question:\\
\emph{``Can we define mathematical measures quantifying useful aspects of creative behavior of 
popular \imgtoimg generation models?''}

We restrict our attention to the aspects of creativity that can be quantified and used to choose a
generative model most suitable for a specific purpose. For example, a generative model suitable for producing images for
a physics textbook would probably have very different creativity-related characteristics than one suitable for 
generating illustrations for a fantasy novel. Using \cite{Hauhio_2024}'s idea of uncontrollability 
based on artifact spaces, as well as the \emph{novel}, \emph{useful}, and 
\emph{typical} criteria laid out by \cite{Peeperkorn_2024}, we propose to characterize outputs of models
according to three criteria, whose formal quantitative definitions we present in the definitions section: 
(1) satisfaction of prompt requirements, (2) cohesion between output artifacts, 
and (3) diversity of output artifacts. We introduce an iterative process involving construction of a \emph{chain} 
for measuring these with a high level of granularity and analyze the results through both statistical and visual means.
The chain construction is inspired by the \emph{telephone} game and includes a repeated generation of new outputs based on
the previous ones. This process is repeated for a predefined number of steps, with the resulting chain analyzed for
the concepts introduced above.
To the best of our knowledge, while previous papers have described aspects of creative behavior, this is the first attempt at mathematically measuring useful aspects of creativity in popular \imgtoimg generation models. 

The chain construction is illustrated in Figure~\ref{fig:flux-strength09-chain-types-examples}, which also shows how
outputs with low creativity scores vs high ones look like. Specifically, the creativity score of the top chain is $0$, 
as it does not satisfy the prompt requirements at all; the creativity score
of the bottom chain is $0.46$, as, though faithful to the prompt, it does not introduce any new elements; 
the creativity score of the middle chain is $0.7$, as it is both faithful to the prompt and introduces new
elements that are cohesive with the ones in the original image.

We show experimental results for evaluating a number of popular image generators with a variety of input temperatures. 
Input temperature, which influences the distribution from which output features are selected, 
has been proposed as a metric for measuring creative behavior in generative models, but this has engendered debate, 
as a high input temperature can easily lead to an unwanted and excessive level of hallucination. 
Our measures capture hallucinations, as well as simply copying the seed image; 
both are ranked low on creativity. 

Due to the lack of space, we only present the summary results and a small number of illustrative images. The code, full experimental results, and full sets of
images and chains can be downloaded from \url{https://figshare.com/s/d88d4966b606163d02fc}.

\section{Background}\label{sec:back}
Here, we provide a detailed breakdown of the concepts and definitions on which we base our experimental setup. 

\subsection{Computational Creativity and Temperature}
A high value of temperature, a parameter used to determine ``the proportion by which the output probability distribution should be flattened'' according to \cite{Peeperkorn_2024}, has been suggested to be a strong influence on creativity, as it increases the chance of more unique, low-probability elements being selected for representation in the model's output \cite{Chen_2023}. This has been described as ``uncontrollability'' by \cite{Hauhio_2024}, and ``hallucination'' in other contexts pertaining to creative behavior in models, such as \cite{Minhyeok_23,Shengqiong_2023}. These studies posit that temperature is a strong influence on creative behavior.

However, other studies argue that using high temperature alone to measure creativity does not allow for any differentiation between (1) high-quality images that use given text or image input prompts as inspiration, and (2) random images produced with little to no influence from the input prompts \cite{Peeperkorn_2024}. While a high temperature makes it likelier for a model to add low-probability elements to its output, there is no guarantee that those elements will be seen as high-quality by the user. 

Given this disagreement, as well as the fact that prominent papers like \cite{Peeperkorn_2024} and \cite{Chen_2023} focus on large language models, we have chosen to probe further into the effects of temperature on \imgtoimg models as a novel contribution to the literary body.

We also note a gap in the literature where \imgtoimg model behavior is concerned. Prior studies on creativity in generative AI models have tended to deal with large language models or \texttoimg generation \cite{Hauhio_2024,Gao_2024}. The creative behavior of \imgtoimg generation, however, is largely unstudied.

\subsection{Defining Facets of Creative Behavior in Generative Vision Models}

We use the term ``artifact'', derived from \cite{Hauhio_2024}'s usage as well as the common linguistic meaning, to refer to a core feature of a text or image, such as ``apple pie'' in the textual prompt ``a slice of apple pie'', or a segment of an image labeled ``pie'' by an object detection tool. Each prompt and output can therefore be understood as a set of artifacts.

\cite{Hauhio_2024} proposes multiple artifact spaces, the relationship between which can be used to classify models' behavior as creative. $C$, referred to as the ``conceptual set'' by \cite{Hauhio_2024}, is the set of artifacts specified in the user's input prompt to a generative model, and $V$ is the set of artifacts a user would find valuable given that this user has a particular goal $\mathbb{R}_A$ in using the generative model. Uncontrollability, in this paper, refers to the idea that a given generative model, when fed $C$, will use a strategy that produces a valuable artifact set $V$ that contains extraneous artifacts not specified in $C$ as well as those in $\mathbb{R}_A$. This expansion of valuable artifacts from strictly the ones referred to in the prompt can, according to this paper, be understood as a basis for creative behavior.

\section{Definitions}\label{sec:def}

Using \cite{Hauhio_2024}'s idea of uncontrollability based on artifact spaces, as well as the \emph{novel}, \emph{useful}, and  \emph{typical} criteria laid out by \cite{Peeperkorn_2024}, we propose to characterize outputs of models
according to three criteria, whose formal quantitative definitions we present below: input prompt \rs, \coh, and \dive. 

\RS, the idea that an output contains (all) the elements specified in the user's input, builds off \cite{Peeperkorn_2024}'s statement that an output must be typical of its class, as well as \cite{Hauhio_2024}'s ``concept set'' $C$. We build \coh from \cite{Peeperkorn_2024}'s assertion that an output must also be \emph{useful} to be considered creative, by working off the idea that an output with elements that are generally more closely related to each other has more use cases than an output with disjoint elements. \coh is also derived from \cite{Boden_1998}'s ``association of ideas'': how well does a generative model build strong associations between artifacts within its output in response to a given prompt? For \dive, we build off \cite{Hauhio_2024}'s idea of valuable artifacts that were not specified by the input prompt, but are added spontaneously by the generative model, as well as \cite{Peeperkorn_2024}'s and \cite{Boden_1998}'s ideas that an output must contain novel elements to be considered creative. Specifically, Boden mentions ``transformational'' creativity as the idea that new elements can be introduced into a space through transformation, a behavior directly linked to the ability of \imgtoimg models to generate novel output elements using the framework of an image input.

Our measure differs from prior diversity metrics like Fr\'{e}chet Inception Distance (FID) \cite{Heusel_2017} in that it solely focuses on individual detectable artifacts within an image, rather than an image's overall structure. Contrary to FID and similar measures, our goal is to optimize the measure for images that differ notably from the 
original image while still containing elements (artifacts) from the original.

In the definitions below we use the following notations: $M$ is the generative model, 
$A_O$ is the artifact set of a given output image $O_M$ (generated by $M$), and
$A_I$ is the artifact set of the input prompt $I$, which can be an image or a text. The \emph{cosine similarity} measure, 
which is used throughout these definitions, is a way of measuring the similarity between the vector projections of two artifacts in semantic embedding space---in other words, a way to mathematically compare the meanings behind textual representations of artifacts. It will be henceforth denoted as the function $\theta$ or as the 
dot product $\cdot$, depending on context. For a set $A$ of artifacts, the embedding of each artifact $a \in A$ 
in semantic space is calculated using the embedding function $emb$, denoted as $emb(a) = \hat{a}$~\cite{Matsuki_2019}.

\begin{definition}[\RS]
    An image is said to be \emph{satisfying requirements} if
    $ A_I \subseteq A_O $. In other words, all artifacts in $I$ are present in $O_M$.
    \label{rs-def}
\end{definition}

\begin{definition}[\Coh]
    \emph{Cohesiveness} $C(O_M)$ of an output image $O_M$ is defined wrt the input $I$. Given the set $A_I$ of the artifacts in $I$ and
    the set $A_O$ of the artifacts in $O_M$, let $P$ be the set of all pairs $(a,b)$, where $a \in A_O \setminus A_I$ and
    $b \in A_I$. Furthermore, let $M_N = \max_{(c,d)\in A_O \setminus A_I}((\hat{c}) \cdot (\hat{d}))$. Then, cohesiveness of $O_M$ is defined as
\[ C(O_M) = \frac{{\sum}_{a:(a,b)\in P}{\max_{b:(a,b)\in P}((\hat{a}) \cdot (\hat{b}))} + M_N}{|P| + 1}. \] 
    \label{coh-def}
\end{definition}
In other words, the cohesiveness of $O_M$ is the mean value of the maximal similarities between new artifacts and seed artifacts and the maximal similarity between new artifacts.

We note that the scalar product $(\hat{a}) \cdot (\hat{b})$ of the representations of $a$ and $b$ in the
semantic embedded space increases with increasing the similarity. As $0 \leq (\hat{a}) \cdot (\hat{b}) \leq 1$, also $0 \leq C(O_M) \leq 1$.
\begin{definition}[\Dive]
    \emph{Diversity} $D(O_M)$ of an output image $O_M$ is defined as the proportion of new artifacts in the image, if $O_M$ contains
    all artifacts of $I$.
    In other words, $D(O_M)$ is $\frac{|A_O \setminus A_I|}{|A_O|}$ if $A_I \subseteq A_O$ and is $0$ otherwise. 
    \label{dive-def}
\end{definition}

While diversity and cohesiveness are not independent concepts, an output image $O_M$ can have 
high diversity and low cohesiveness if it contains many new artifacts with low similarity to the original ones, 
and it can also have high cohesiveness and low diversity if
there are very few new artifacts and they are very similar to the original ones.

\section{Methodology}
In this section, we outline our methodology for quantitative evaluation of creativity of image generation models.

\subsection{Chain Construction}

As \cite{Hauhio_2024} points out, image generation models are often used iteratively, with users modifying their input prompts based on the artifacts present in the generated output. In this paper, we simulate this iterative approach, which allows us to measure the afore-discussed aspects of creative behavior 
in a way that is more pertinent to real-world use-cases. To our knowledge, an iterative approach like ours has not been utilized for experimentation in prior computational creativity studies.

\begin{figure}[!htb]
\centering
    \includegraphics[width=0.5\columnwidth]{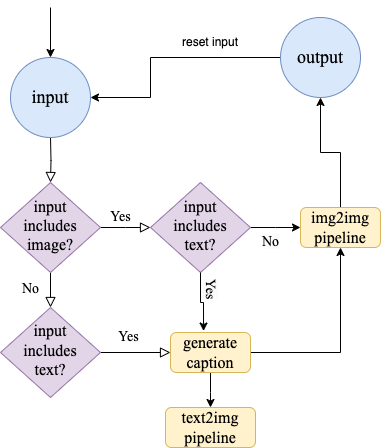}
\caption{The flowchart depicting a single\\step in the chain process}
\label{fig:chain-diagram}
\end{figure}

We base our approach on the well-known game \emph{telephone} (or \emph{Chinese whispers}). 
As in this game, each chain in our experiment begins with a ``ground truth'', or seed image. 
This image is then processed into an input for an image generation model, and the output of this is  
and fed back into the image generation model; this process repeats for $x$ iterations, 
where $x$ is a parameter chosen by the user. 
One step of the process is illustrated in the flowchart in Figure \ref{fig:chain-diagram}.

The specific mechanisms of the input processing mentioned above are determined by the type of chain. Our main focus is \imgtoimg generative models, hence our experimental setup evaluates how strongly an image input influences aspects of creativity, compared to textual input. We implement three types of chains, 
as shown in Table \ref{tab:chain-types}. Generative image models can be used with text input prompts, image input prompts, or a combination of both, which is reflected in our experimental setup. For chains with text input, \texttoimg, we auto-generate a caption for the prior image 
in the chain and feed that into the image generation model in the next step. For chains with image input (\imgtoimg), we pass 
the prior image in the chain directly into the image generation model. 

\begin{table}[H]
    \small
    \caption{Chain Types in Experimental Process}
    \label{tab:chain-types}
    \centering
    \begin{tabularx}{\columnwidth}{X|X|X}
        Chain Type & Image Seed &   Text Seed \\
        \midrule\midrule
        \imonly\ & Previous image  &   None \\
        \hline
        \caponly\ & None &   Caption of Previous Image \\
        \hline
        \imcap & Previous Image &   Caption of Previous Image \\
    \end{tabularx}
\end{table}

Based on the earlier discussion of the effects of temperature and uncontrollability on creative behavior, we consider the temperature as a parameter
of the image inputs to the \imgtoimg pipelines. Each of the \imgtoimg pipeline types is executed with multiple levels of temperature for the image input, 
to determine whether different temperatures have statistically significant effects on the \rs, \coh, \dive, and overall \creativity of the outputs 
of a given generation model.

\subsection{Measures for Creativity}

Using the conceptual definitions of \rs, \coh, and \dive provided earlier, we can define mathematical measures for each of these aspects in the context of the chains, so that we can quantify the creative behaviors exhibited by the generator used within a given chain.

The \satfact measure provides the backbone for the other two, as we want to focus on images that exhibit a higher-than-random level of \rs for the input prompt. We expand and quantify Definition \ref{rs-def} to produce this measure for a chain of images rather than a single comparison between two images. For a given chain, we quantify \satfact as the 
\emph{normalized longest unbroken sequence} of the chain wherein, at each step, the seed artifact set, or 
a set of artifacts that approximate the seed artifact set within a cosine similarity threshold $t$, 
is a subset of the artifact set of the image generated for that step.

We then multiply this proportion of the chain by the average $\theta$ score for the set of pairs consisting of one artifact from the seed set and its closest match in the generated image's artifact set, to account for the fact that the generated image may not be perfectly faithful. For example if the seed artifact set is [``apple pie'',  ``horse''], and the generated image for which \satfact is calculated has the artifact set [``apple cake'', ``horse''], the proportion of the chain that satisfies the input prompt as described in the previous paragraph would be multiplied by 
\[ \frac{\theta(\mbox{``apple pie''}, \mbox{``apple cake''}) + \theta(\mbox{``horse''}, \mbox{``horse''})}{2} \] 
to differentiate it between a generated image that contains the exact seed artifact set.

The later in the chain the seed artifacts or their approximations still appear, the higher the \satfact value is going to 
be, although it also depends on how closely the artifacts in the generated image approximate the chain image.

Let $l$ be the maximal length of the chain, $O_n$ the image produced at step $n$ of the chain, where $1 \leq n \leq l$, 
$A_S$ the textual set of artifacts present in the chain's seed image, $A_n$ the textual set of artifacts present in $O_n$, and $t$ the threshold of approximation between seed and generated image artifacts:

Let $K$, for $0 \leq K \leq l$, be the number of consecutive steps from the beginning of a chain for which the following holds: for all $k \leq K$ and $a_s \in A_S$, there exists $a_k \in O_k$, where
\begin{equation}
(\hat{a_k})\cdot (\hat{a_s})\ \geq\ t.
\label{rs-equation-part1}
\end{equation}
A chain's \SatFact is then defined as
\begin{equation}
    RS =  \frac{K}{l} \times mean((\hat{a_k})\cdot (\hat{a_s})).
\label{rs-equation-part2}
\end{equation}

In other words, we extract artifacts from the generated image in each step of the chain. 
We then represent each artifact in the current image as a vector in semantic embedding space, and calculate the
cosine similarity between
these embeddings and the embeddings of the artifacts in the seed image. $K$ is the index of the last step, where 
where all artifacts fall within the similarity threshold $t$ of their counterparts in the seed image.
The chain's \SatFact is defined as the product of $K$ normalized by the overall chain length $l$ with the mean value
of the similarities between artifacts of the generated images and their closest counterparts in the seed image.

Consider the image $O_K$ (the $K$-st image in the chain). As it is the last image in the generated chain that satisfies
Equation~\ref{rs-equation-part1}, we use it to analyze cohesion and diversity, rather than calculating those for the
whole chain.
Let $A_K$ be the artifact set of $O_K$, and $A_S$ once again represent the artifact set of the seed image.
A chain's \CohFact $B_R$ (using $B$ for bond, to differentiate from $CR$, the creativity score as defined later) can be calculated using Definition \ref{coh-def}. We set $P$ to be the set of all pairs $(a, b)$, 
where $a \in A_K \setminus A_S$ and $b \in A_K \cap A_S$:

\begin{equation}
    m_1 = {\sum}_{a:(a,b)\in P}({\max_{b:(a,b)\in P}((\hat{a}) \cdot (\hat{b}))})
    \label{cohesion-factor-m1}
\end{equation}
Let $M_N$ be as in Definition \ref{coh-def}, with $A_O = A_K$ and $A_I = A_S$. Then, $B_R$ is defined as:
\begin{equation}
B_R = \begin{cases}
    \frac{M_N + \sum_{m \in m_1} m}{|m_1| + 1} \text{ if } |A_K| \geq 2\\
    0 \text{ if } |A_K| < 2
\end{cases}
\label{cohfact-equation}
\end{equation}

Breaking this equation down step-by-step, the \cohfact is calculated by first creating a set of all 
\emph{new} artifacts in $O_K$, that is, $A_K \setminus A_S$. 
Then we calculate the similarity of these artifacts to the artifacts brought from the seed image,
in Equation~\ref{cohesion-factor-m1}. $M_N$ stands for the similarity between the new artifacts themselves.
The $B_R$ score is the mean value of these, where the similarity between the new artifacts has a lower weight than the
similarity between the new artifacts and the seed image ones.
We account for instances where $A_K$ has less than $2$ artifacts by automatically setting $B_R$ to be $0$, 
as cohesion requires at least two objects. 

The diversity score captures the fraction of $A_K$ consisting of labels that are not found in $A_S$, 
indicating how many extraneous objects were added to $O_K$, and is based on Definition \ref{dive-def}:
\begin{equation}
D_R = \begin{cases}
    \frac{|A_K \setminus A_S|}{|A_K|} \text{ if } A_S \subseteq A_K\\
    \text{ else } 0.
\end{cases}
\label{divefact-equation}
\end{equation}

Each of these quantify individual behaviors for a given chain. 
In order to measure overall creativity, we combine these measures to an overall \emph{creativity ranking} score, 
$CR$, where $0 \leq CR \leq 1$. For a given chain:
\begin{equation}
    CR = RS\times\frac{B_R + D_R}{2}.
\label{cr-equation}
\end{equation}
The chain under evaluation is a parameter to all scores we introduce in this section; it is omitted for brevity.

Intuitively, the creativity score can be higher than $0$ only if the generated images satisfy the prompt requirements; this
prevents us from awarding a high creativity score to hallucinating models. On the other hand, a mechanistic copying of
the seed image is not creative either. Indeed, while $RS=1$ for a mechanistic copy, both $B_R$ and $D_R$ are $0$, hence
resulting in the creativity score $0$ as well. For models that satisfy the prompt requirements, while adding new elements
to the generated images, both the diversity between the new elements and their cohesion with the seed image elements, as
well as between themselves, affect the creativity score.

If we are given a set $X$ of generative models, we can use the means of $RS$, $B_R$, $D_R$, and $CR$ across the total number of chains for all models in $X$ to rank them in terms of specific aspects of creative behavior as per our earlier discussion. Of course, since creativity is inherently subjective and may vary across users and contexts, this ranking may not be universally applicable. Extensions or modifications to these definitions can be explored to better fit specific domains or to align with different interpretations of creativity.

\section{Evaluation}

\subsection{Experimental Setup}

For our experiments we selected sets of simple seed images focusing on a single, easy-to-detect subject, as these
would be the easiest to measure \rs, \coh, and \dive, since $A_S$ would be $1$. Since many object detectors are trained on datasets that prominently feature food images, we pulled our $999$ seed images from the 
publicly-available \texttt{Food-101} dataset~\cite{Bossard_2014}. We selected three food categories at random: 
``apple pie'', ``donuts'', and ``pizza'', and randomly selected $333$ images from each. Input prompts $I$ were
constructed from each of these seed images.

We evaluated our measures on three popular open-source image generation models from 
Hugging Face: \sd3 \cite{Esser_2024}, \kand \cite{Razzhigaev_2023}, and \flux \cite{BFL_2023}.
The nature of the input prompt $I$ depends on the type of chain as shown in Table \ref{tab:chain-types}. 
For the former two chain types, we pass the seed image into the image generation model using the appropriate 
\imgtoimg pipeline. For the latter two chain types, we generate a text caption for the seed image using \kosmos 
model from Hugging Face \cite{Peng_2023}. The rest of the chain is constructed using the same principle: 
for \imgtoimg chains, the next entry is produced by passing in an $I$ consisting of the previous image, 
while for \texttoimg chains $I$ contains the \kosmos caption for the previous image. 
For chains with a combination of both caption and image input, both the previous image and its \kosmos 
caption are passed in. 
In our experiments, we set the maximal length of the chain to $10$ to avoid computational explosion, while still producing meaningful results. 

For temperature, we used the \emph{strength} parameter as described in the Hugging Face \imgtoimg pipeline 
documentation \cite{HLiu_2023}. The documentation
claims that strength influences the creativity of the pipeline's output, a claim we aim to test with our experiments. 
Functionally, strength determines the amount of noise added to the seed image during the generative process, meaning that a lower value of strength adds less noise and therefore produces an output that more closely resembles the seed image, while a higher value of strength adds more noise to the input image and thus produces an output that is visually farther from the seed image. We executed both the \imonly\ and \imcap\ pipelines for all three \imgtoimg models with three different strength levels: $0.3$, $0.6$, and $0.9$, thus encompassing an equally spaced range from low to high. Since strength controls the number of generation steps in \imgtoimg pipelines and is not present in \texttoimg pipelines, for the sake of accurate comparison we executed each of the \caponly\ pipelines with three generation step counts: $15$, corresponding to the number of generation steps used by the strength of $0.3$ pipelines; $30$, for the strength of $0.6$; and $45$, corresponding to the strength $0.9$. In the rest of this paper, we use ``temperature'' and ``strength'' interchangeably.


To compute the measures of \rs, \coh, \dive, and creativity ranking defined earlier, we used two object detectors, 
\dino and \detr, to extract artifacts in textual form from images in the chain, where the initial image is the seed image,
and for $A_S$ we simply use a text representation of the subject of the seed image: apple pie, donut, or pizza. 
For word embedding and semantic similarity calculations between the artifacts, we used the \spacy library. We selected $0.65$ as the value of $t$ for \satfact, after running preliminary tests to account for incorrect object detection results.

We used \emph{paired T-tests} to determine any statistically significant differences between results for different strengths, chain types, and image generation models, on the same dataset of $999$ images. 
The reasoning behind this choice is
that finding statistically significant differences between different chain configurations 
would yield information about the impact of image vs text inputs on our creativity measures, 
as well as allow us to compare the relative performance of different generative models.
We used \scipy's \texttt{ttest\_rel} function to perform these tests.

\subsection{Presentation of Results}

\begin{figure}[!htb]
    \small
     \begin{subfigure}{\columnwidth}
    \centering
    \includegraphics[width=0.08\columnwidth]{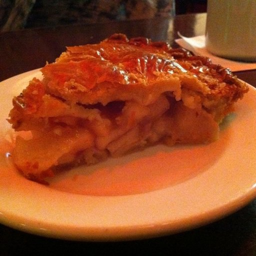}
    \includegraphics[width=0.08\columnwidth]{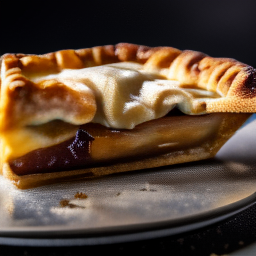}
    \includegraphics[width=0.08\columnwidth]{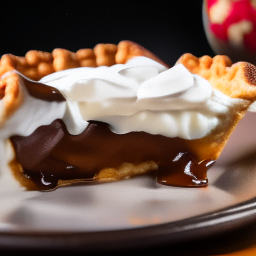}
    \includegraphics[width=0.08\columnwidth]{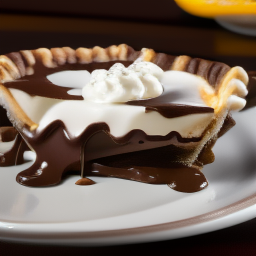}
    \includegraphics[width=0.08\columnwidth]{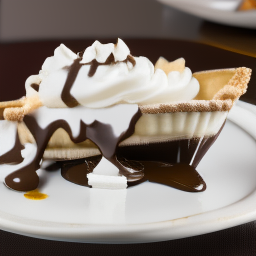}
    \includegraphics[width=0.08\columnwidth]{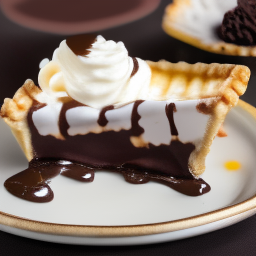}
    \includegraphics[width=0.08\columnwidth]{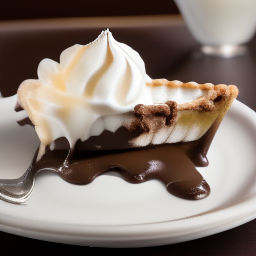}
    \includegraphics[width=0.08\columnwidth]{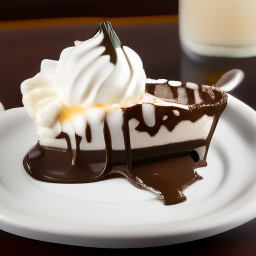}
    \includegraphics[width=0.08\columnwidth]{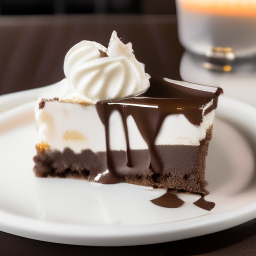}
    \includegraphics[width=0.08\columnwidth]{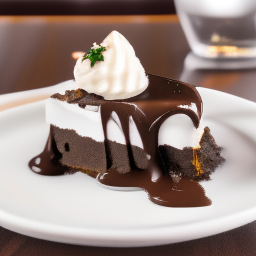}
    \includegraphics[width=0.08\columnwidth]{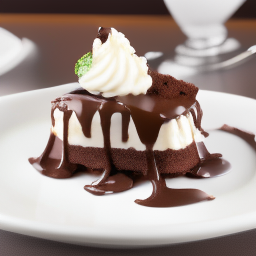}
    \caption{At strength 0.3.}\label{subfig:a}
    \end{subfigure}
    
    \begin{subfigure}{\columnwidth}
    \centering
    \includegraphics[width=0.08\columnwidth]{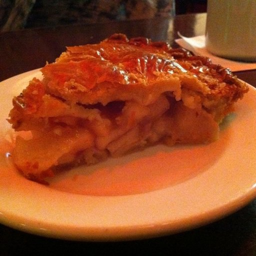}
    \includegraphics[width=0.08\columnwidth]{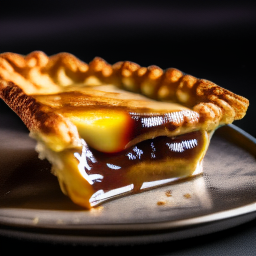}
    \includegraphics[width=0.08\columnwidth]{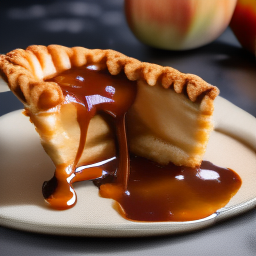}
    \includegraphics[width=0.08\columnwidth]{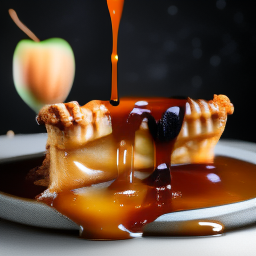}
    \includegraphics[width=0.08\columnwidth]{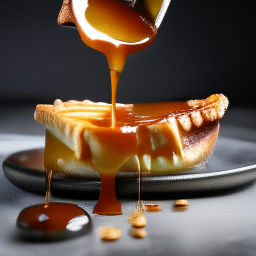}
    \includegraphics[width=0.08\columnwidth]{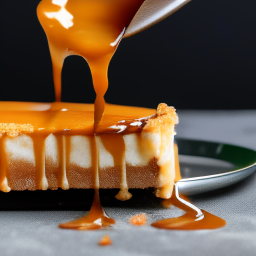}
    \includegraphics[width=0.08\columnwidth]{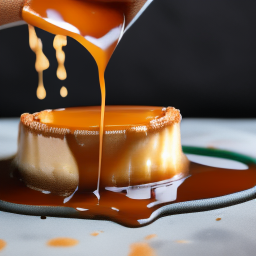}
    \includegraphics[width=0.08\columnwidth]{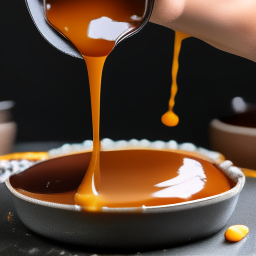}
    \includegraphics[width=0.08\columnwidth]{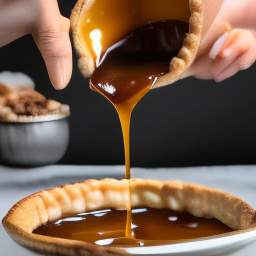}
    \includegraphics[width=0.08\columnwidth]{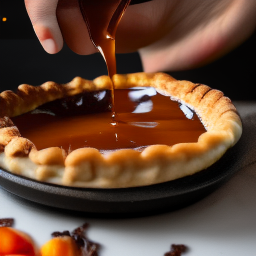}
    \caption{At strength 0.6.}\label{subfig:b}
    \end{subfigure}

    \begin{subfigure}{\columnwidth}
    \centering
    \includegraphics[width=0.08\columnwidth]{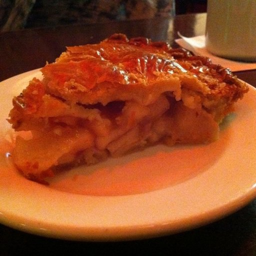}
    \includegraphics[width=0.08\columnwidth]{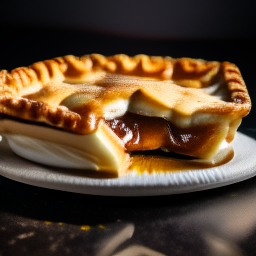}
    \includegraphics[width=0.08\columnwidth]{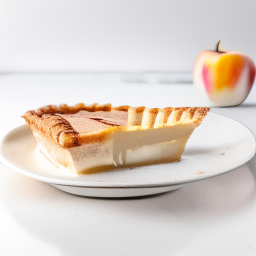}
    \includegraphics[width=0.08\columnwidth]{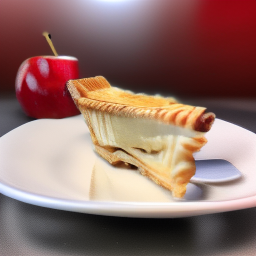}
    \includegraphics[width=0.08\columnwidth]{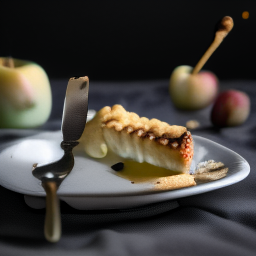}
    \includegraphics[width=0.08\columnwidth]{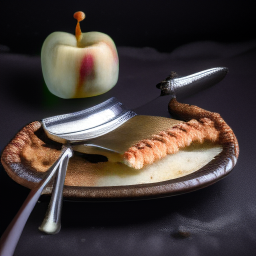}
    \includegraphics[width=0.08\columnwidth]{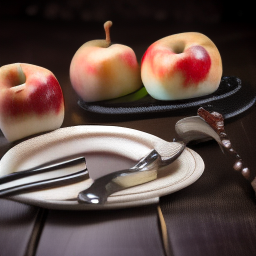}
    \includegraphics[width=0.08\columnwidth]{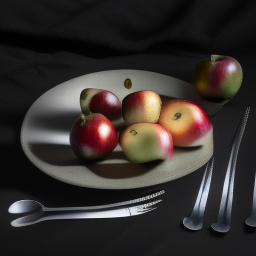}
    \includegraphics[width=0.08\columnwidth]{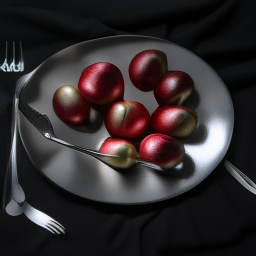}
    \includegraphics[width=0.08\columnwidth]{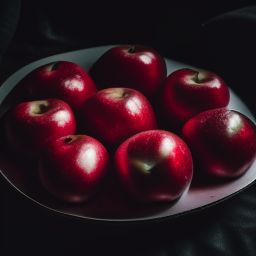}
    \includegraphics[width=0.08\columnwidth]{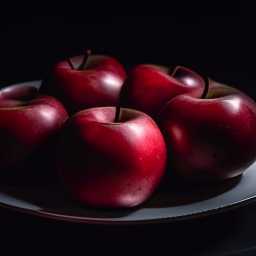}
    \caption{At strength 0.9.}\label{subfig:c}
    \end{subfigure}
    \caption{Example \imcap\ \kand chains for various strengths, generated from a single seed image \texttt{apple\_pie\_184.png}. 
    From left to right: the seed image, followed by steps 1-$n$ of the generated chain where $n$ is the last image that satisfies the prompt requirements.}
    \label{fig:kand-apple-pie-example}
\end{figure}

The values of $RS$, $B_R$, and $D_R$ were calculated for the chains generated for each of the $27$ possible combinations
of image generation model, strength, and chain type. \dino and \detr were used on each generated image to extract artifact labels from them in text form.

Figure \ref{fig:kand-apple-pie-example} is an example of how our creativity measures manifest visually in a set of \imcap\ \kand chains at different strengths. The chain pictured in \Cref{subfig:a} was generated with strength $0.3$. 
The \satfact value of this chain is $0.68$, meaning that while the chain remains unbroken due to the average 
cosine similarity staying above the threshold of $0.65$, the closest artifact found in the last image, ``cake'', 
only has a similarity of $0.68$ to the seed artifact, ``apple pie''. 
Accordingly, the chain values of \cohfact, \divefact, and creativity ranking as calculated from the last image 
in this chain are $B_R = 0.3; D_R = 0.67; CR = 0.48$, indicating a medium creativity score, due to relatively high diversity (``syrup'', ``dining table'', and ``cup'' were added in) and low to medium cohesion between the artifacts. 
The \satfact value of the chain shown in \Cref{subfig:b}, generated with strength $0.6$, is $0.9$, 
indicating that the subject remains ``apple pie'' until step $9$ out of $10$. Using the $9$th image 
(the rightmost pictured), the chain's scores are $B_R = 0.45; D_R = 0.75; CR = 0.54$, indicating a higher
creativity than the previous chain, due to an increased number of added labels, particularly a person's hand pouring syrup. The \satfact of the chain shown in \Cref{subfig:c} is also $0.9$, with the closest artifact match being ``apple'', 
but because no other labels were identified by either \dino or \detr in the image, $D_R = 0.0$. A single subject yields a \cohfact value of $0$, hence $B_R = 0.0$ as well. Therefore, $CR = 0.0$, indicating a chain that adequately satisfies the prompt requirements but, unlike the first two, does not contain any novel elements. 

Given the standard threshold of $p < 0.05$ to indicate a statistically significant difference between sets of results, almost all the models showed statistically significant differences. The tables and figures 
below show averages and comparisons of these measures across \imcap\ chains for each combination of image generation models with different strength. Values of $p$ that are extremely close to $0$ are represented as $0+\epsilon$. 
Note that the mean values are rounded to $2$ decimal points, meaning that some means that look very close may in fact 
belong to statistically different datasets.

\begin{table*}[!htb]
    \small
    \caption{Paired T-Test results for $CR$ in \imcap\ chains with strength 0.3}
    \label{tab:img-cap-gen-comparisons-0.3}
    \centering
    \begin{tabularx}{\textwidth}{X|X|X|X|X}
                model 1 & model 2 & mean 1 & mean 2 & P-value \\
        \midrule\midrule
        \flux & \kand & 0.18 & 0.19 & 0.5007\\
        \midrule
        \flux & \sd & 0.18 &  0.1 & $0+\epsilon$\\
        \midrule
        \kand & \sd & 0.19 &  0.1 & $0+\epsilon$\\
    \end{tabularx}
\end{table*}

\begin{table*}[!htb]
    \small
    \caption{Paired T-Test results for $CR$ in \imcap\ chains with strength 0.6}
    \label{tab:img-cap-gen-comparisons-0.6}
    \centering
    \begin{tabularx}{\textwidth}{X|X|X|X|X}
                model 1 & model 2 & mean 1 & mean 2 & P-value \\
        \midrule\midrule
        \flux & \kand & 0.2 & 0.17 & 0.0220\\
        \midrule
        \flux & \sd & 0.2 &  0.1 & $0+\epsilon$\\
        \midrule
        \kand & \sd & 0.17 &  0.1 & $0+\epsilon$\\
    \end{tabularx}
\end{table*}

\begin{table*}[!htb]
    \small
    \caption{Paired T-Test results for $CR$ in \imcap\ chains with strength 0.9}
    \label{tab:img-cap-gen-comparisons-0.9}
    \centering
    \begin{tabularx}{\textwidth}{X|X|X|X|X}
                model 1 & model 2 & mean 1 & mean 2 & P-value \\
        \midrule\midrule
        \flux & \kand & 0.22 & 0.19 & 0.0011\\
        \midrule
        \flux & \sd & 0.22 &  0.13 & $0+\epsilon$\\
        \midrule
        \kand & \sd & 0.19 &  0.13 & $0+\epsilon$\\
    \end{tabularx}
\end{table*}

In summary, as shown by Figures \ref{fig:kand-apple-pie-example}, \ref{fig:apple-pie-example-imonly}, \ref{fig:apple-pie-example-imcap}, and Tables \ref{tab:img-cap-gen-comparisons-0.3}, \ref{tab:img-cap-gen-comparisons-0.6}, \ref{tab:img-cap-gen-comparisons-0.9}, \ref{tab:caponly-imcap-comparisons-flux}, \ref{tab:caponly-imcap-comparisons-kand}, and \ref{tab:caponly-imcap-comparisons-sd}, applying our defined measures to the chains generated from our experimental process yields statistical significant differences between the various combinations we used. In the
next section we analyze what those differences mean in the context of our definitions of \RS, \Coh, and \Dive.  

\subsection{Analysis}

Our first observation is that the average creativity ranking, $CR$, across all the models regardless of temperature was relatively low, ranging from $0.0$ to $0.24$. Manual examination showed that these low scores were due to multiple
factors: early deviation from the prompt requirements, satisfaction of prompt requirements but low diversity, or the generation of unclassifiable objects such as glitchy patterns or irregular geometric shapes. 
None of the image generation models we tested performed too well in terms of creativity. 
However, our statistical tests and manual observations demonstrated significant differences 
in the scores between chain types and models, as detailed below.

As can be seen in Tables~\ref{tab:caponly-imcap-comparisons-flux}, \ref{tab:caponly-imcap-comparisons-kand}, and \ref{tab:caponly-imcap-comparisons-sd}, the paired T-tests demonstrate significant statistical differences
between the results in all the \imcap\ and \caponly\ chains, supporting the claim that input images have
a measurable impact on model outputs.
Furthermore, for \flux and \sd, there are statistically significant differences in $CR$ values for \imcap\ chains between strength $0.3$ and $0.9$, with $CR$ values for higher-temperature chains being higher on average. This lends support to our original hypothesis that a higher image input temperature increases the creativity of an \imgtoimg model's output.

\begin{table}[!htb]
    \small
    \caption{Paired T-Test Results for $CR$ in \flux chains}
    \label{tab:caponly-imcap-comparisons-flux}
    \centering
    \begin{tabularx}{\textwidth}{c |p{1.8cm}|p{1.8cm}| c }
        strength & \imcap\ mean & \caponly\ mean & P-value  \\
        \midrule\midrule
        0.3 & 0.18 & 0.24 & $0+\epsilon$\\
        \midrule
        0.6 & 0.2 & 0.27 & $0+\epsilon$\\
        \midrule
        0.9 & 0.22 & 0.27 & $0+\epsilon$\\
    \end{tabularx}
\end{table}

\begin{table}[!htb]
    \small
    \caption{Paired T-Test results for $CR$ in \kand chains}
    \label{tab:caponly-imcap-comparisons-kand}
    \centering
    \begin{tabularx}{\textwidth}{c |p{1.8cm}|p{1.8cm}| c }
        strength & \imcap\ mean & \caponly\ mean & P-value  \\
        \midrule\midrule
        0.3 & 0.19 & 0.23 & $0+\epsilon$\\
        \midrule
        0.6 & 0.17 & 0.25 & $0+\epsilon$\\
        \midrule
        0.9 & 0.19 & 0.23 & $0+\epsilon$\\
    \end{tabularx}
\end{table}

\begin{table}[!htb]
    \small
    \caption{Paired T-Test results for $CR$ in \sd chains}
    \label{tab:caponly-imcap-comparisons-sd}
    \centering
    \begin{tabularx}{\textwidth}{c |p{1.8cm}|p{1.8cm}| c }
        strength & \imcap\ mean & \caponly\ mean & P-value  \\
        \midrule\midrule
        0.3 & 0.1 & 0.19 & $0\ +\ \epsilon$\\
        \midrule
        0.6 & 0.1 & 0.21 & $0\ +\ \epsilon$\\
        \midrule
        0.9 & 0.13 & 0.2 & $0\ +\ \epsilon$\\
    \end{tabularx}
\end{table}

\begin{figure}[!htb]
    \begin{subfigure}{\columnwidth}
    \centering
    \includegraphics[width=0.2\columnwidth]{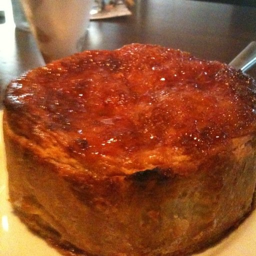}
    \caption{Seed image \texttt{apple\_pie\_027.png}}\label{subfig:applepie}
    \end{subfigure}
    \hfill
    \begin{subfigure}{\columnwidth}
    \centering
    \includegraphics[width=0.15\columnwidth]{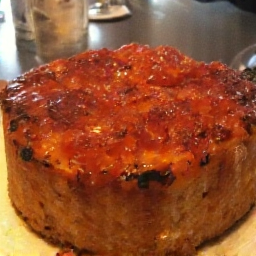}
    \includegraphics[width=0.15\columnwidth]{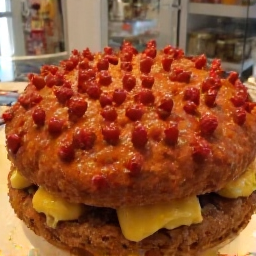}
    \includegraphics[width=0.15\columnwidth]{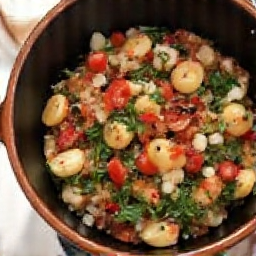}
    \end{subfigure}
    \begin{subfigure}{\columnwidth}
    \centering
    \includegraphics[width=0.15\columnwidth]{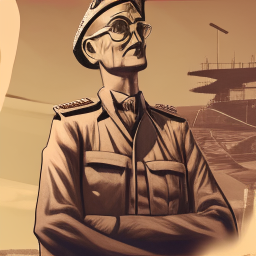}
    \includegraphics[width=0.15\columnwidth]{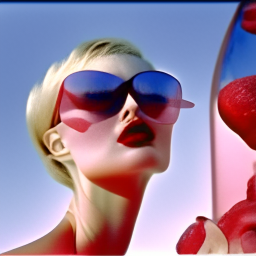}
    \includegraphics[width=0.15\columnwidth]{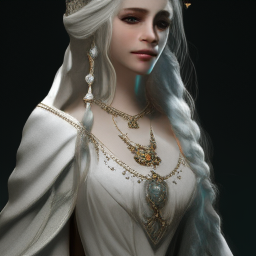}
    \end{subfigure}
    \begin{subfigure}{\columnwidth}
    \centering
    \includegraphics[width=0.15\columnwidth]{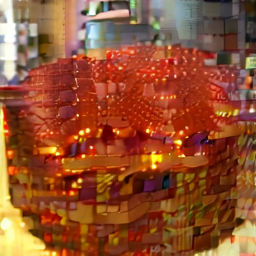}
    \includegraphics[width=0.15\columnwidth]{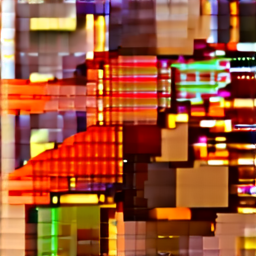}
    \includegraphics[width=0.15\columnwidth]{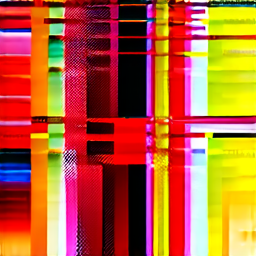}
    \end{subfigure}
    \caption{Example images generated at step $1$ of their respective \imonly\ chains, all seeded with the same image. \flux in the top row, \kand in the middle row, and \sd in the bottom row; strength $0.3$ in the leftmost column, $0.6$ in the middle column, and $0.9$ in the rightmost column.}
    \label{fig:apple-pie-example-imonly}
\end{figure}

Notably, all of the \imonly\ chains demonstrated statistically significant differences from the other types of chains, 
with the average $CR$ values for \imonly\ chains being significantly lower than the \imcap\ and \caponly\ 
averages across all generation models and strengths. Only \flux \imonly\ chains for strength $0.3$ and $0.6$ 
have average $CR$ values above $0.0$ (see Figure \ref{fig:apple-pie-example-imonly} for an illustration).

The data in Tables \ref{tab:caponly-imcap-comparisons-flux}, \ref{tab:caponly-imcap-comparisons-kand}, and \ref{tab:caponly-imcap-comparisons-sd} and the statistically significant differences between \imcap\ chains of different temperatures
further confirm our hypothesis that the temperature of image inputs directly influences the overall creativity of the 
output. However, the low average creativity scores of \imonly\ chains also indicate that captions are important for image generation models, even the \imgtoimg ones we tested. The analysis below takes into account
the granular measures \SatFact ($RS$), \CohFact ($B_R$), and \DiveFact ($D_R$), focusing on \imonly\ and \imcap\ chains.

The values of $RS$ in \flux decrease sharply as strength increases, with $RS=0.66$ for strength $0.3$ and $RS=0.0$
for strength $0.9$. For other too models, \sd3 and \kand, average $RS$ is close to $0$ across all strengths.

For $RS$ we first looked at \imonly\ chains, observing that in \flux, $RS$ decreases drastically as strength increases, starting at 0.66 for strength = 0.3 and ending up at 0.0 for strength = 0.9; while for the other two models, average $RS$ values are extremely low, approaching 0, across all strengths. This observation is supported by the visual examination of
randomly sampled images created within \imonly\ chains for each generator and is illustrated by 
Figure \ref{fig:apple-pie-example-imonly}, which shows a series of images generated from the same seed image, \texttt{apple\_pie\_027.png} in \Cref{subfig:applepie}, at the first step of each of their respective \imonly\ chains, with
varying strengths. The extremely low requirement satisfaction is evident here: only the \flux chain at strength $0.3$ 
shows an image resembling an apple pie. 

\imcap\ chains, by contrast, have high average $RS$ values across all three models, in the $0.7-0.85$ range. 
These results indicate that the models we tested rely heavily on textual captions for correct and creative image
generation, with the partial exception of \flux exhibiting some ability to generate faithful images from image input.
At the same time, \imcap\ chains for all three models demonstrate a 
statistically significant difference in $RS$ value between strengths $0.3$ and $0.9$, 
indicating that image input strength affects the model's ability to satisfy its original prompt even with textual captions. 

\begin{figure}[!htb]
    \begin{subfigure}{\columnwidth}
    \centering
    \includegraphics[width=0.15\columnwidth]{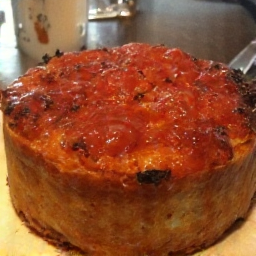}
    \includegraphics[width=0.15\columnwidth]{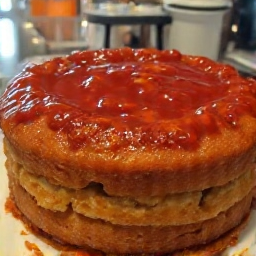}
    \includegraphics[width=0.15\columnwidth]{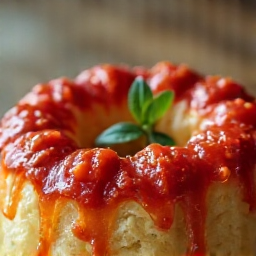}
    \end{subfigure}
    \begin{subfigure}{\columnwidth}
    \centering
    \includegraphics[width=0.15\columnwidth]{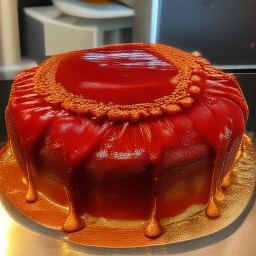}
    \includegraphics[width=0.15\columnwidth]{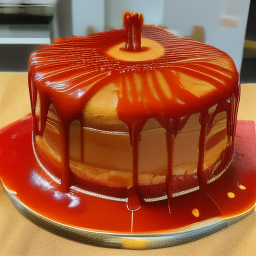}
    \includegraphics[width=0.15\columnwidth]{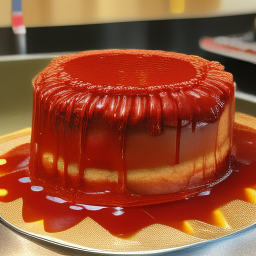}
    \end{subfigure}
    \begin{subfigure}{\columnwidth}
    \centering
    \includegraphics[width=0.15\columnwidth]{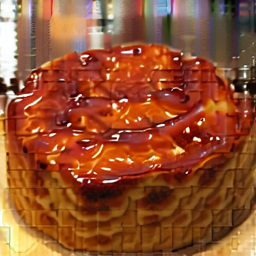}
    \includegraphics[width=0.15\columnwidth]{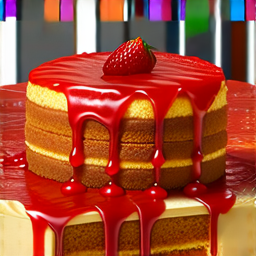}
    \includegraphics[width=0.15\columnwidth]{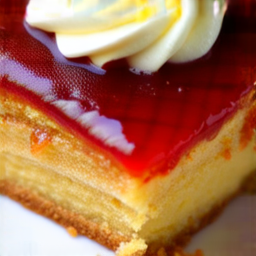}
    \end{subfigure}
    \caption{Example images generated at step 1 of their respective \imcap chains, all seeded with the same image. \flux in the top row, \kand in the middle row, and \sd in the bottom row; strength $0.3$ in the leftmost column, $0.6$ in the middle column, and $0.9$ in the rightmost column.}
    \label{fig:apple-pie-example-imcap}
\end{figure}

This claim is further supported by the fact that all three models show statistically significant differences 
between \imcap\ and \imonly\ chains for strengths $0.6$ and $0.9$, with \sd showing differences between strengths 
$0.3$ and $0.6$ as well. For \imcap\ chains created with \flux and \sd, strength $0.9$ results in the highest average $RS$ 
values. The most significant effect of strength is observed in \sd chains, and the least significant one in \flux. 
Figure \ref{fig:apple-pie-example-imcap} shows the first image in each chain generated from the same seed image 
in \Cref{subfig:applepie}, also organized by generator and strength level. Unlike the 
\imonly\ chains, all of these initial images satisfy the prompt requirement by including the artifact ``apple pie''.

Adding to our observation that image input strength has a direct relationship with our creativity measures, average $B_R$ scores for \imcap\ chains across all three models show a statistically significant increase between chains generated with strength $0.3$ and those generated with strength $0.9$. \imonly\ chains, on the other hand, only show statistically significant differences for \sd, with all three models hovering around an average $B_R$ of $0.14$ regardless of strength. This disparity indicates that textual input is particularly important for cohesion.

$D_R$, like $B_R$, is informed by $RS$: the \divefact for a given chain is automatically set to 0.0 if the prompt requirements are not satisfied within the chain. \imcap\ chains for both \flux and \kand display no statistically significant differences between each other, and have average values around 0.25, but \sd differs from both, with a significantly lower average $D_R$ around 0.13. In contrast, all three models produced statistically significant differences between their \imonly\ chains, but only \flux chains have average $D_R$ values significantly higher than 0.0. Unlike the other measures, diversity does not have a straightforwardly direct relationship with image input strength, but seems to heavily rely on textual input.

To summarize, our results show that image input strength has a measurable effect on the \satfact, \cohfact, 
and overall \emph{creativity ranking} of the outputs produced by a given \imgtoimg generation model, while diversity appears to be less influenced by it. However, in order to produce outputs that score well in terms of creativity ranking, 
textual input is also crucial for all three models; out of the three that we tested, \flux seems to be the best at 
interpreting image-only inputs through a creative lens.

\section{Limitations}

Apart from computational resources, a significant limitation of our approach is its reliance on 
existing popular object detection and word similarity calculation tools. These models are prone to occasionally making decisions that do not seem intuitive to humans, such as placing two seemingly unrelated words close together in semantic embedding space, or labeling one object with a similar but incorrect designation (e.g. labeling a slice of apple pie as a slice of cake). There is currently no way around this limitation, but as the outputs of such tools more closely approach human-level understanding, the quality of our measures will improve accordingly. Additionally, in order to improve the visual accuracy and explainability of our proposed measures, future research could incorporate human observation as a factor in validating the scores output.

We also acknowledge that the simplicity of our seed image dataset, although easy to experiment on and ideal for observing changes, could impact creativity scores. Seed images with a large artifact set may potentially increase $D_R$, for example, or make it harder for generated images to satisfy prompt requirements. However, exploring the creativity across 
diverse domains and more complex datasets subjects is a crucial next step.

\section{Conclusions and Future Research}
In this paper, we suggest a set of measures to quantitatively evaluate creativity in \imgtoimg generative models. Our experimental results visually and statistically demonstrate that
the image input temperature (also referred to as ``strength'') plays a statistically significant and measurable role in manifestations of creative behavior of \imgtoimg generative models. Moreover, 
textual input is also crucial for creative behavior; lack of textual input leads to generated images that bear very
little resemblance to the input image. 

We can, therefore, answer the research question posed in the introduction affirmatively: our metrics provide a fine-grained score of creative behavior that conforms relatively well to human intuition and can be used to determine which model
and at which strength is most suitable for a particular task. 

In the future, some existing metrics, such as LPIPS \cite{Isola_Zhang_2018} for requirement satisfaction, could be 
incorporated into our creativity ranking to improve the granularity of our measures.
Further directions for this research include exploring why some generative models score differently, and whether these metrics can help tailor future model architecture and training techniques to specific purposes.

\clearpage
\bibliographystyle{apalike}
\bibliography{references}

\end{document}